# Counterfactual Probabilities: Computational Methods, Bounds and Applications

**Alexander Balke**
Cognitive Systems Laboratory
University of California
Los Angeles, CA 90024
*balke@cs.ucla.edu*

**Judea Pearl**
Cognitive Systems Laboratory
University of California
Los Angeles, CA 90024
*judea@cs.ucla.edu*

## Abstract

Evaluation of counterfactual queries (e.g., "If $A$ were true, would $C$ have been true?") is important to fault diagnosis, planning, and determination of liability. In this paper we present methods for computing the probabilities of such queries using the formulation proposed in [Balke and Pearl, 1994], where the antecedent of the query is interpreted as an external action that forces the proposition $A$ to be true. When a prior probability is available on the causal mechanisms governing the domain, counterfactual probabilities can be evaluated precisely. However, when causal knowledge is specified as conditional probabilities on the observables, only bounds can computed. This paper develops techniques for evaluating these bounds, and demonstrates their use in two applications: (1) the determination of treatment efficacy from studies in which subjects may choose their own treatment, and (2) the determination of liability in product-safety litigation.

## 1 INTRODUCTION

A counterfactual sentence has the form

If $A$ were true, then $C$ would have been true where $A$, the counterfactual antecedent, specifies an event that is contrary to one's real-world observations, and $C$, the counterfactual consequent, specifies a result that is expected to hold in the alternative world where the antecedent is true. A typical instance is "If Oswald were not to have shot Kennedy, then Kennedy would still be alive" which presumes the factual knowledge of Oswald's shooting Kennedy, contrary to the antecedent of the sentence.

Because of the tight connection between counterfactuals and causal influences, any algorithm for computing solutions to counterfactual queries must rely heavily on causal knowledge of the domain. This leads naturally to the use of probabilistic causal networks, since these networks combine causal and probabilistic knowledge and permit reasoning from causes to effects as well as, conversely, from effects to causes.

To emphasize the causal character of counterfactuals, we adopt the interpretation in [Pearl, 1993b], according to which a counterfactual sentence "If it were $A$, then $B$ would have been" states that $B$ would prevail if $A$ were forced to be true by some unspecified action that is exogenous to the other relationships considered in the analysis.

Causal theories specified in functional form (as in [Pearl and Verma, 1991, Druzdzel and Simon, 1993, Poole, 1993]) are sufficient for evaluating counterfactual queries, whereas the causal information embedded in Bayesian networks is not sufficient for the task. Every Bayes network can be represented by several functional specifications, each yielding different evaluations of a counterfactual. The problem is that, deciding what factual information deserves undoing (by the antecedent of the query) requires a model of temporal persistence, and, as noted in [Pearl, 1993c], such a model is not part of static Bayesian networks. Functional specifications, however, implicitly contain the needed temporal persistence information.

Consider an example with two variables $A$ and $B$, representing Ann and Bob's attendance, respectively, at a party ($A = a_1$ when Ann is at the party, $A = a_0$ otherwise; $B = b_1$ when Bob is at the party, $B = b_0$ otherwise), and it is believed that Ann's attendance has a causal influence on Bob's attendance, shown by the arrow $A \rightarrow B$). Assume that previous behavior shows $P(b_1|a_1) = 0.9$ and $P(b_0|a_0) = 0.9$. We observe that Bob and Ann are absent from the party and we wonder whether Bob would be there if Ann were there. The answer depends on the mechanism that accounts for the 10% exception in Bob's behavior. If the reason Bob occasionally misses parties (when Ann goes) is that he is unable to attend (e.g., being sick or having to finish a paper for UAI), then the answer to our query would be 90%. However, if the only reason for Bob's occasional absence (when Ann goes) is that he becomes angry with Ann (in which case he does exactly the opposite of what she does), then the answer to our query is 100%, because Ann and Bob's current absence from the party proves that Bob is not angry.

Thus, we see that the information contained in the conditional probabilities on the observed variables is insufficient for answering counterfactual queries uniquely; some information about the mechanisms responsible for these probabilities is needed as well. Still, when only a probabilistic model is given, informative bounds on the counterfactual probabilities can often be derived, and this paper provides a general framework for evaluating these bounds.

The next section will introduce concise notation for expressing counterfactual queries. Section 3.2 will derive a general expression for counterfactual probabilities in terms of a functional specification. Section 3.3 will present a general procedure for evaluating bounds on counterfactual probabilities when only a probabilistic specification is supplied. Section 4 will apply this procedure for evaluating bounds on treatment effects in partial compliance studies, while Section 5 will demonstrate the use of this procedure in product liability litigation.

## 2  NOTATION

Let the set of variables describing the world be designated by $X = \{X_1, X_2, \ldots, X_n\}$. As part of the complete specification of a counterfactual query, there are real-world observations that make up the background context. These observed values will be represented in the standard form $x_1, x_2, \ldots, x_n$. In addition, we must represent the value of the variables in the counterfactual world. To distinguish between $x_i$ and the value of $X_i$ in the counterfactual world, we will denote the latter with an asterisk; thus, the value of $X_i$ in the counterfactual world will be represented by $x_i^*$. We will also need a notation to distinguish between events that might be true in the counterfactual world and those referenced explicitly in the counterfactual antecedent. The latter are interpreted as being forced to the counterfactual value by an external action, which will be denoted by a hat (e.g., $\hat{x}$).

Thus, a typical counterfactual query will have the form "What is $P(c^*|\hat{a}^*, a, b)$?" to be read as "Given that we have observed $A = a$ and $B = b$ in the real world, if $A$ were $\hat{a}^*$, then what is the probability that $C$ would have been $c^*$?"

## 3  BOUNDS ON COUNTERFACTUALS

In [Balke and Pearl, 1994], an algorithm was presented for evaluating the unique quantitative solutions to counterfactual queries when a functional model is given. In this section we briefly describe the form of the functional model using *response-function* variables and how the solution is evaluated uniquely. Then we deal with probabilistic specifications and show how bounds can be obtained by optimizing the solution above over all functional models consistent with the probabilistic specification.

### 3.1  FUNCTIONAL MODELS

For the previously described party example, a functional specification models the influence of Ann's attendance ($A$) on Bob's attendance ($B$) by a deterministic function

$$b = F_b(a, \epsilon_b)$$

where $\epsilon_b$ stands for all unknown factors that may influence $B$ and the prior probability distribution $P(\epsilon_b)$ quantifies the likelihood of such factors. For example, whether Bob has been grounded by his parents and whether Bob is angry at Ann could make up two possible components of $\epsilon_b$. Given a specific value for $\epsilon_b$, $B$ becomes a deterministic function of $A$; hence, each value in $\epsilon_b$'s domain specifies a *response function* that maps each value of $A$ to some value in $B$'s domain. In general, the domain for $\epsilon_b$ could contain many components, but it can always be replaced by an equivalent variable that is minimal, by partitioning the domain into equivalence regions, each corresponding to a single response function [Pearl, 1993a]. Formally, these equivalence classes can be characterized as a function $r_b : \text{dom}(\epsilon_b) \to \mathbf{N}$, as follows:

$$r_b(\epsilon_b) = \begin{cases} 0 & \text{if } F_b(a_0, \epsilon_b) = 0 \ \& \ F_b(a_1, \epsilon_b) = 0 \\ 1 & \text{if } F_b(a_0, \epsilon_b) = 0 \ \& \ F_b(a_1, \epsilon_b) = 1 \\ 2 & \text{if } F_b(a_0, \epsilon_b) = 1 \ \& \ F_b(a_1, \epsilon_b) = 0 \\ 3 & \text{if } F_b(a_0, \epsilon_b) = 1 \ \& \ F_b(a_1, \epsilon_b) = 1 \end{cases}$$

Obviously, $r_b$ can be regarded as a random variable that takes on as many values as there are functions between $A$ and $B$. We will refer to this domain-minimal variable as a *response-function variable*. $r_b$ is closely related to the *potential response variables* in Rubin's model of counterfactuals [Rubin, 1974], which was introduced to facilitate causal inference in statistical analysis [Balke and Pearl, 1993].

For this example, the response-function variable for $B$ has a four-valued domain $r_b \in \{0, 1, 2, 3\}$ with the following functional specification:

$$b = f_b(a, r_b) = h_{b,r_b}(a) \qquad (1)$$

where the mappings defined by each response function $h_{b,r_b}(a)$ are given by

$$h_{b,0}(a) = b_0 \quad , \quad h_{b,1}(a) = \begin{cases} b_0 & \text{if } a = a_0 \\ b_1 & \text{if } a = a_1 \end{cases}$$

$$h_{b,3}(a) = b_1 \quad , \quad h_{b,2}(a) = \begin{cases} b_1 & \text{if } a = a_0 \\ b_0 & \text{if } a = a_1 \end{cases}$$

The response-function variable for $A$ has a two-valued domain $r_a \in \{0, 1\}$ with the functional specification:

$$a = f_a(r_a) = h_{a,r_a}()$$

where

$$h_{a,0}() = a_0 \quad , \quad h_{a,1}() = a_1$$

The prior probability on the response functions $P(r_b)$ and $P(r_a)$ in conjunction with $f_b(a, r_b)$ and $f_a(r_a)$ fully parameterizes the model.

For each observable variable $X_i$, there will be a function that maps the value of $X_i$'s observable causal influences $\text{pa}(X_i)$ and $X_i$'s response-function variable $r_{x_i}$ to the value of $X_i$

$$x_i = f_{x_i}(\text{pa}(x_i), r_{x_i})$$

If the model is *complete* (such as the functional model described in [Pearl and Verma, 1991]), all response functions will be mutually independent, and each will be characterized by a prior probability $P(r_{x_i})$. However, when some variables are left out of the analysis, the response functions of the remaining variables $(x_1, \ldots, x_n)$ may be dependent and, in principle, a joint probability $P(r_{x_1}, \ldots, r_{x_n})$ would be required. In practice, only local dependencies will be needed.

If one assumes that two variables $A$ and $B$ are dependent via some exogenous common cause, then we create an edge between $r_a$ and $r_b$ and specify the joint distribution $P(r_a, r_b)$. This treatment of latent variables will be utilized in the applications discussed in Sections 4 and 5.

### 3.2 FUNCTIONAL EXPRESSION

We now derive an expression for $P(c^*|\hat{a}^*, o)$ in terms of the underlying functional model.

The connection between the factual and counterfactual worlds is discussed in [Balke and Pearl, 1994] where it is argued that the response-function variables should assume the same values in both worlds. For the party example, this invariance allows the response function variables $r_a$ and $r_b$ to be shared between the networks corresponding to the two worlds (see Figure 1).

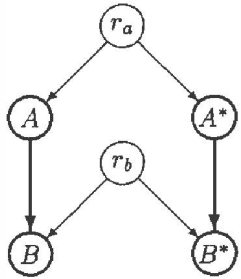

Figure 1: *Factual $(A, B)$ and counterfactual $(A^*, B^*)$ worlds for the functional analysis of the structure $A \rightarrow B$. The response-function variables $r_a$ and $r_b$ (summarizing all exogenous influences on $A$ and $B$) attain the same value in the real and counterfactual worlds.*

Let $\mathbf{r} = (r_{x_1}, r_{x_2}, \ldots, r_{x_n})$ represent the set of response-function variables for all the variables in the model. Given the value of $\mathbf{r}$, all variables $X_i \in X$ are functionally determined according to the recursive function:

$$\begin{aligned} x_i &= f_{x_i}(\mathbf{r}) \\ &= f_{x_i}(f_{u_1}(\mathbf{r}), f_{u_2}(\mathbf{r}), \ldots, f_{u_k}(\mathbf{r}), r_{x_i}) \end{aligned}$$

where $\text{pa}(X_i) = \{U_1, U_2, \ldots, U_k\} \subset X$ are the causal influences of $X_i$ in the model.

If a set of variables $A \subset X$ in the model are externally forced to the value $\hat{a}$, then according to the action-based semantics of [Pearl, 1993a], the recursive function becomes

$$\begin{aligned} x_i &= f_{x_i}^{\hat{a}}(\mathbf{r}) \\ &= \begin{cases} \hat{x}_i & \text{if } X_i \in A \\ f_{x_i}(r_{x_i}) & \text{if } X_i \notin A \text{ and } \text{pa}(X_i) = \emptyset \\ f_{x_i}(f_{u_1}^{\hat{a}}(\mathbf{r}), f_{u_2}^{\hat{a}}(\mathbf{r}), \ldots, f_{u_k}^{\hat{a}}(\mathbf{r}), r_{x_i}) & \text{otherwise} \end{cases} \end{aligned}$$

The counterfactual probability $P(c^*|\hat{a}^*, o)$ may be rewritten

$$P(c^*|\hat{a}^*, o) = \frac{P(c^*, o|\hat{a}^*)}{P(o|\hat{a}^*)}$$

Since an action can only affect its descendants in the graph [Pearl, 1994] we have $P(o|\hat{a}) = P(o)$ which is readily computed from the probabilistic specification.

$P(c^*, o|\hat{a}^*)$ may be evaluated in terms of the functional model by summing the probabilities of the response-function configurations which are consistent with the arguments $(c^*, \hat{a}^*, o)$. Formally,

$$P(c^*, o|\hat{a}^*) = \sum_{\mathbf{r} \in R} P(\mathbf{r})$$

where

$$R = \{\mathbf{r}|\forall_{x_i \in o}[x_i = f_{x_i}(\mathbf{r})] \text{ and } \forall_{x_j^* \in c^*}[x_j^* = f_{x_j}^{\hat{a}}(\mathbf{r})]\}$$

Hence, the counterfactual probability may be written in terms of the structure $\{\text{pa}(x_i)\}$ and parameters $P(\mathbf{r})$ of the functional model:

$$P(c^*|\hat{a}^*, o) = \frac{\sum_{\mathbf{r} \in R} P(\mathbf{r})}{P(o)} \quad (2)$$

In the next section this expression will be optimized under the constraints imposed by the probabilistic specification.

### 3.3 CONSTRAINTS AND OPTIMIZATION

The probabilistic specification $P(x_i|\text{pa}(x_i))$ for a complete model imposes a set of constraints on $P(r_{x_i})$ of the form

$$P(x_i|\text{pa}(x_i)) = \sum_{r_{x_i}} P(r_{x_i}) t(r_{x_i}; x_i, \text{pa}(x_i)) \quad (3)$$

where the characteristic function $t$ indicates which values of $r_{x_i}$ map the particular value of $X_i$'s causal influences $(\text{pa}(x_i))$ to the specific value of $X_i$ $(x_i)$, i.e.

$$t(r_{x_i}; x_i, \text{pa}(x_i)) = \begin{cases} 1 & \text{if } x_i = f_{x_i}(\text{pa}(x_i), r_{x_i}) \\ 0 & \text{otherwise} \end{cases}$$

For an incomplete model, if $X_i$ and $X_j$ are assumed to have an exogenous common cause, then the common constraint for these two variables will be given instead by

$$P(x_i, x_j|\text{pa}(x_i) - \{x_j\}, \text{pa}(x_j) - \{x_i\}) = \quad (4)$$
$$\sum_{r_{x_i}, r_{x_j}} P(r_{x_i}, r_{x_j}) t(r_{x_i}; x_i, \text{pa}(x_i)) t(r_{x_j}; x_j, \text{pa}(x_j))$$

Note that the constraints in Eq. (4) are linear in $P(r_{x_i}, r_{x_j})$.

For example, in the party story (which is complete with two binary variables $A$ and $B$) the constraints are given by

$$P(b_1|a_0) = P(r_b=2) + P(r_b=3)$$
$$P(b_1|a_1) = P(r_b=1) + P(r_b=3)$$
$$P(a_1) = P(r_a=1)$$

Additional subjective constraints may also be imposed on the underlying functional model. For example, we may subjectively believe that Bob is never spiteful against Ann, which can be simply written $P(r_b=2) = 0$ and added to the existing set of constraints.

Given the entire set of linear constraints and the objective function from Eq. (2), the bounds may be evaluated using techniques for optimizing non-linear objective functions under linear constraints [Scales, 1985]. In general, the optimization procedure may converge to a local minima/maxima which would produce false bounds. If the objective is to prove that the counterfactual probability falls within a certain range, care must be taken to ensure that global optima are found.

If the objective function given by Eq. (2) is linear, the minimum/maximum may be determined using linear programming techniques. In this case, when the problem size is small enough, we may also derive symbolic bounds to the counterfactual probability in terms of the probabilistic specification. This is accomplished by tracking the conditions that lead to the various decisions in the Simplex Tableau algorithm. This procedure generates a decision tree where each leaf node contains a symbolic solution [Balke and Pearl, 1993].

## 4 APPLICATION TO CLINICAL TRIALS WITH IMPERFECT COMPLIANCE

Consider an experimental study where random assignment has taken place but compliance is not perfect (i.e., the treatment received differs from that assigned). It is well known that under such conditions a bias may be introduced, in the sense that the true causal effect of the treatment may deviate substantially from the causal effect computed by simply comparing subjects receiving the treatment with those not receiving the treatment. Because the subjects who did not comply with the assignment may be precisely those who would have responded adversely (positively) to the treatment, the actual effect of the treatment, when applied uniformly to the population, might be substantially less (more) effective than the study reveals.

In an attempt to avert this bias, economists have devised correctional formulas based on an "instrumental variables" model ([Bowden and Turkington, 1984]) which, in general, do not hold outside the linear regression model. A recent analysis by [Efron and Feldman, 1991] departs from the linear regression model, but still makes restrictive commitments to a particular mode of interaction between compliance and response. [Robins, 1989] and [Manski, 1990] derived nonparametric bounds on treatment effects using different techniques; however their bounds are not tight. [Holland, 1988] has given a general formulation of the problem (which he called "encouragement design") in terms of Rubin's model of causal effect and has outlined its relation to path analysis and structural equations models. [Angrist et al., 1993], also invoking Rubin's model, have identified a set of assumptions under which the "Instrumental Variable" formula is valid for certain subpopulations. These subpopulations cannot be identified from empirical observation alone, and the need remains to devise alternative, assumption-free formulas for assessing the effect of treatment over the population as a whole. In this section, we derive bounds on the average treatment effect that rely solely on observed quantities and are universal, that is, valid no matter what model actually governs the interactions between compliance and response.

The canonical partial-compliance setting can be graphically modeled as shown in Figure 2.

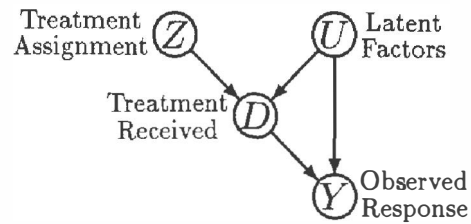

Figure 2: *Graphical representation of causal dependencies in a randomized clinical trial with partial compliance.*

We assume that $Z$, $D$, and $Y$ are observed binary variables where $Z$ represents the (randomized) treatment assignment, $D$ is the treatment actually received, and $Y$ is the observed response. $U$ represents all factors, both observed and unobserved, that may influence the outcome $Y$ and the treatment $D$. To facilitate the notation, we let $z$, $d$, and $y$ represent, respectively, the values taken by the variables $Z$, $D$, and $Y$, with the following interpretation: $z \in \{z_0, z_1\}$, $z_1$ asserts that treatment has been assigned ($z_0$, its negation); $d \in \{d_0, d_1\}$, $d_1$ asserts that treatment has been administered ($d_0$, its negation); and $y \in \{y_0, y_1\}$, $y_1$ asserts a positive observed response ($y_0$, its negation). The domain of $U$ remains unspecified and may, in general, combine the spaces of several random variables, both discrete and continuous.

The graphical model reflects two assumptions of independence:

1. The treatment assignment does not influence Y directly, but only through the actual treatment D, that is,

$$Z \perp\!\!\!\perp Y \mid \{D, U\} \tag{5}$$

In practice, any direct effect $Z$ might have on $Y$ would be adjusted for through the use of a placebo.

2. $Z$ and $U$ are marginally independent, that is, $Z \perp\!\!\!\perp U$. This independence is partly ensured through the randomization of $Z$, which rules out a common cause for both $Z$ and $U$. The absence of a direct path from $Z$ to $U$ represents the assumption that a person's disposition to comply with or deviate from a given assignment is not in itself affected by the assignment; any such effect can be viewed as part of the disposition.

These assumptions impose on the joint distribution[1] the decomposition

$$P(y,d,z,u) = P(y|d,u)\,P(d|z,u)\,P(z)\,P(u) \quad (6)$$

which, of course, cannot be observed directly because $U$ is a latent variable. However, the marginal distribution $P(y,d,z)$ and, in particular, the conditional distributions $P(y,d|z), z \in \{z_0, z_1\}$, are observed, and the challenge is to assess the causal effect of $D$ on $Y$ from these distributions.[2]

In addition to the independence assumption above, the causal model of Figure 2 reflects claims about the behavior of the population under external interventions. In particular, it reflects the assumption that $P(y|d,u)$ is a stable quantity: the probability that an individual with characteristics $U = u$ given treatment $D = d$ will respond with $Y = y$ remains the same, regardless of how the treatment was selected — be it by choice or by policy. Therefore, if we wish to predict the distribution of $Y$ under a condition where the treatment $D$ is applied uniformly to the population, we should calculate

$$P(y^*|\hat{d}^*) = \sum_u P(y|d,u)P(u) \quad (7)$$

Likewise, if we are interested in estimating the average *change* in $Y$ due to treatment, we define the average *causal effect*, $\mathrm{ACE}(D \to Y)$ ([Holland, 1988]), as

$$\mathrm{ACE}(D \to Y) = P(y_1^*|\hat{d}_1^*) - P(y_1^*|\hat{d}_0^*) \quad (8)$$

The task of causal inference is then to estimate or bound the expression in Eq. (8), given the observed probabilities $P(y,d|z_0)$ and $P(y,d|z_1)$. This may be accomplished by following the procedure detailed in Section 3.3 where the objective function to be optimized is the difference between the two counterfactual probabilities on the right-hand side of Eq. (8).

First, the functional model corresponding to the probabilistic model of Figure 2 must be specified. For each of the observable variables in the model ($Z$, $D$, and $Y$), we define the corresponding response-function variables ($r_z$, $r_d$, and $r_y$, respectively).

Figure 3 shows the graphical representation of the resulting functional model. Because $D$ and $Y$ are assumed to be influenced by an unobservable common cause, the response-function variables $r_d$ and $r_y$ are connected by an edge.

The states of the variables $r_d$ and $r_y$ have the following interpretations:

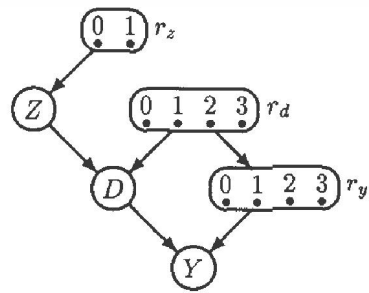

Figure 3: *A structure equivalent to that of Figure 1 but employing response-function variables $r_z$, $r_d$ and $r_y$.*

$D$ is a deterministic function of the variable $Z$ and $r_d \in \{0,1,2,3\}$:

$$d = f_d(z, r_d) = h_{d,r_d}(z)$$

where

$$h_{d,0}(z) = d_0 \quad, \quad h_{d,1}(z) = \begin{cases} d_0 & \text{if } z = z_0 \\ d_1 & \text{if } z = z_1 \end{cases}$$

$$h_{d,3}(z) = d_1 \quad, \quad h_{d,2}(z) = \begin{cases} d_1 & \text{if } z = z_0 \\ d_0 & \text{if } z = z_1 \end{cases}$$

Similarly, $Y$ is a deterministic function of $D$ and $r_y \in \{0,1,2,3\}$:

$$y = f_y(d, r_y) = h_{y,r_y}(d) \quad (9)$$

where

$$h_{y,0}(d) = y_0 \quad, \quad h_{y,1}(d) = \begin{cases} y_0 & \text{if } d = d_0 \\ y_1 & \text{if } d = d_1 \end{cases}$$

$$h_{y,3}(d) = y_1 \quad, \quad h_{y,2}(d) = \begin{cases} y_1 & \text{if } d = d_0 \\ y_0 & \text{if } d = d_1 \end{cases}$$

The correspondence between the states of variables $r_d$ and $r_y$ and the potential response vectors in the Rubin's model [Rosenbaum and Rubin, 1983] is rather transparent: each state corresponds to a counterfactual statement specifying how a unit in the population (e.g., a person) would have reacted to any given input. For example, $r_d = 1$ represents units with perfect compliance, while $r_d = 2$ represents units with perfect defiance. Similarly, $r_y = 1$ represents units with perfect response to treatment, while $r_y = 0$ represents units with no response ($y = y_0$) regardless of treatment. The counterfactual variables $Y_1$ and $Y_0$ usually invoked in Rubin's model can be obtained from $r_y$ as follows:

$$Y_1 = \{Y \text{ if } D = d_1\} = \begin{cases} 1 & \text{if } r_y = 1 \text{ or } r_y = 3 \\ 0 & \text{otherwise} \end{cases}$$

---

[1] We take the liberty of denoting the prior distribution of $U$ by $P(u)$, even though $U$ may consist of continuous variables.

[2] In practice, of course, only a finite sample of $P(y,d|z)$ will be observed, but since our task is one of identification, not estimation, we make the large-sample assumption and consider $P(y,d|z)$ as given.

$$Y_0 = \{Y \text{ if } D = d_0\} = \begin{cases} 1 & \text{if } r_y = 2 \text{ or } r_y = 3 \\ 0 & \text{otherwise} \end{cases}$$

In general, treatment response and compliance attitudes may not be independent, hence the arrow $r_d \to r_y$ in Figure 3. The joint distribution over $r_d \times r_y$ requires 15 independent parameters, and these parameters are sufficient for specifying the model of Figure 3, $P(y, d, z, r_d, r_y) = P(y|d, r_y)P(d|r_d, z)P(z)P(r_d, r_y)$, because $Y$ and $D$ stand in functional relation to their parents in the graph. The causal effect of the treatment can now be obtained directly from Eqs. (7) and (9) according to Eq. (2), giving

$$P(y_1^*|\hat{d}_1^*) = P(r_y=1) + P(r_y=3) \quad (10)$$
$$P(y_1^*|\hat{d}_0^*) = P(r_y=2) + P(r_y=3) \quad (11)$$

and

$$\text{ACE}(D \to Y) = P(r_y=1) - P(r_y=2) \quad (12)$$

### 4.1 LINEAR PROGRAMMING FORMULATION

In this section we will explicate the relationship between the parameters of the observed distribution $P(y, d|z)$ and the parameters of the joint distribution $P(r_d, r_y)$ of the response functions. This will lead directly to the linear constraints needed for minimizing/maximizing $\text{ACE}(D \to Y)$ given the observation $P(y, d|z)$.

The conditional distribution $P(y, d|z)$ over the observable variables is fully specified by eight parameters, which will be notated as follows:

$$p_{00.0} = P(y_0, d_0|z_0) \quad p_{00.1} = P(y_0, d_0|z_1)$$
$$p_{01.0} = P(y_0, d_1|z_0) \quad p_{01.1} = P(y_0, d_1|z_1)$$
$$p_{10.0} = P(y_1, d_0|z_0) \quad p_{10.1} = P(y_1, d_0|z_1)$$
$$p_{11.0} = P(y_1, d_1|z_0) \quad p_{11.1} = P(y_1, d_1|z_1)$$

The probabilistic constraints

$$\sum_{n=00}^{11} p_{n.0} = 1 \quad \sum_{n=00}^{11} p_{n.1} = 1 \quad (13)$$

further imply that $\vec{p} = (p_{00.0}, \ldots, p_{11.1})$ can be specified by a point in six-dimensional space. This space will be referred to as $P$.

The joint probability over $r_d \times r_y$, $P(r_d, r_y)$, has 16 parameters and completely specifies the population under study. These parameters will be notated as

$$q_{jk} = P(r_d=j, r_y=k)$$

where $j, k \in \{0, 1, 2, 3\}$. The probabilistic constraint

$$\sum_{j=0}^{3}\sum_{k=0}^{3} q_{jk} = 1$$

implies that $\vec{q}$ specifies a point in 15-dimensional space. This space will be referred to as $Q$.

Eq. (12) can now be rewritten as a linear combination of the $Q$ parameters:

$$\text{ACE}(D \to Y) = \quad (14)$$
$$q_{01} + q_{11} + q_{21} + q_{31} - q_{02} - q_{12} - q_{22} - q_{32}$$

Applying Eqs. (3) and (4) we can write the constraints which reflect the direct linear transformation from a point $\vec{q}$ in $Q$ space to the corresponding point $\vec{p}$ in the observation space $P$:

$$p_{00.0} = q_{00} + q_{01} + q_{10} + q_{11}$$
$$p_{01.0} = q_{20} + q_{22} + q_{30} + q_{32}$$
$$p_{10.0} = q_{02} + q_{03} + q_{12} + q_{13}$$
$$p_{11.0} = q_{21} + q_{23} + q_{31} + q_{33}$$

$$p_{00.1} = q_{00} + q_{01} + q_{20} + q_{21}$$
$$p_{01.1} = q_{10} + q_{12} + q_{30} + q_{32}$$
$$p_{10.1} = q_{02} + q_{03} + q_{22} + q_{23}$$
$$p_{11.1} = q_{11} + q_{13} + q_{31} + q_{33}$$

which will be written in matrix form, $\vec{p} = \bar{P}\vec{q}$.

Given a point $\vec{p}$ in $P$ space, the strict lower bound on $\text{ACE}(D \to Y)$ can be determined by solving the following linear programming problem:

Minimize: $q_{01} + q_{11} + q_{21} + q_{31} - q_{02} - q_{12} - q_{22} - q_{32}$

Subject to:

$$\sum_{j=0}^{3}\sum_{k=0}^{3} q_{jk} = 1$$
$$\bar{P}\vec{q} = \vec{p} \quad (15)$$
$$q_{jk} \geq 0 \text{ for } j, k \in \{0, 1, 2, 3\}$$

However, for problems of this size, the procedure may be used for deriving symbolic expressions as well, leading to the following lower bound on the treatment effect

$$\text{ACE}(D \to Y) \geq \quad (16)$$

$$\max \begin{cases} p_{11.1} + p_{00.0} - 1 \\ p_{11.0} + p_{00.1} - 1 \\ p_{11.0} - p_{11.1} - p_{10.1} - p_{01.0} - p_{10.0} \\ p_{11.1} - p_{11.0} - p_{10.0} - p_{01.1} - p_{10.1} \\ -p_{01.1} - p_{10.1} \\ -p_{01.0} - p_{10.0} \\ p_{00.1} - p_{01.1} - p_{10.1} - p_{01.0} - p_{00.0} \\ p_{00.0} - p_{01.0} - p_{10.0} - p_{01.1} - p_{00.1} \end{cases}$$

Similarly, the upper bound is given by

$$\text{ACE}(D \to Y) \leq$$

$$\min \begin{cases} 1 - p_{01.1} - p_{10.0} \\ 1 - p_{01.0} - p_{10.1} \\ -p_{01.0} + p_{01.1} + p_{00.1} + p_{11.0} + p_{00.0} \\ -p_{01.1} + p_{11.1} + p_{00.1} + p_{01.0} + p_{00.0} \\ p_{11.1} + p_{00.1} \\ p_{11.0} + p_{00.0} \\ -p_{10.1} + p_{11.1} + p_{00.1} + p_{11.0} + p_{10.0} \\ -p_{10.0} + p_{11.0} + p_{00.0} + p_{11.1} + p_{10.1} \end{cases}$$

We may also derive bounds on the treatment responses under the condition where treatment is uniformly applied to the population by optimizing Eqs. (10) and (11) individually (under the same linear constraints). The resulting bounds are:

$$\max \left\{ \begin{array}{c} p_{10.0} + p_{11.0} - p_{00.1} - p_{11.1} \\ p_{10.1} \\ p_{10.0} \\ p_{01.0} + p_{10.0} - p_{00.1} - p_{01.1} \end{array} \right\}$$
$$\leq P(y_1^* | \hat{d}_0^*) \leq$$
$$\min \left\{ \begin{array}{c} p_{01.0} + p_{10.0} + p_{10.1} + p_{11.1} \\ 1 - p_{00.1} \\ 1 - p_{00.0} \\ p_{10.0} + p_{11.0} + p_{01.1} + p_{10.1} \end{array} \right\}$$

and

$$\max \left\{ \begin{array}{c} p_{11.0} \\ p_{11.1} \\ -p_{00.0} - p_{01.0} + p_{00.1} + p_{11.1} \\ -p_{01.0} - p_{10.0} + p_{10.1} + p_{11.1} \end{array} \right\}$$
$$\leq P(y_1^* | \hat{d}_1^*) \leq$$
$$\min \left\{ \begin{array}{c} 1 - p_{01.1} \\ 1 - p_{01.0} \\ p_{00.0} + p_{11.0} + p_{10.1} + p_{11.1} \\ p_{10.0} + p_{11.0} + p_{00.1} + p_{11.1} \end{array} \right\}$$

These bounds improve upon the results of [Manski, 1990]. In addition, one can prove that these are the tightest possible assumption-free bounds.

Examples and additional results regarding bounds on treatment effects in partial compliance studies are presented in [Balke and Pearl, 1993].

## 5 APPLICATIONS TO LIABILITY JUDGMENT

Evaluation of counterfactual probabilities could be enlightening in some legal cases in which a plaintiff claims that a defendant's actions were responsible for the plaintiff's misfortune. Improper rulings can easily be issued without an adequate treatment of counterfactuals. Consider the following hypothetical and fictitious case study, especially crafted to accentuate the disparity between different methods of analysis.

The marketer of PeptAid (antacid medication) randomly mailed out product samples to 10% of the households in the city of Stress, California. In a follow-up study, researchers determined for each individual whether they received the PeptAid sample, whether they consumed PeptAid, and whether they developed peptic ulcers in the following month.

The causal structure which describes the influences in this scenario is identical to the partial-compliance model given by Figure 2, where $z_1$ asserts that PeptAid was received from the marketer; $d_1$ asserts that PeptAid was consumed; and $y_1$ asserts that peptic ulceration occurred. The data showed the following distribution:

$$P(z_1) = 0.1$$

$$P(y_0, d_0 | z_0) = 0.32 \qquad P(y_0, d_0 | z_1) = 0.02$$
$$P(y_0, d_1 | z_0) = 0.32 \qquad P(y_0, d_1 | z_1) = 0.17$$
$$P(y_1, d_0 | z_0) = 0.04 \qquad P(y_1, d_0 | z_1) = 0.67$$
$$P(y_1, d_1 | z_0) = 0.32 \qquad P(y_1, d_1 | z_1) = 0.14$$

This data indicates a high-correlation between those individuals who consumed PeptAid and those who developed peptic ulcers in the following month

$$P(y_1 | d_1) = 0.50 \qquad P(y_1 | d_0) = 0.26$$

In addition, the intent-to-treat analysis showed that those individuals who received the PeptAid samples had a 45% greater chance of developing peptic ulcers

$$P(y_1 | z_1) = 0.81 \qquad P(y_1 | z_0) = 0.36$$

The plaintiff (Mr. Smith), having heard of the study, litigated against both the marketing firm and the PeptAid producer. The plaintiff's attorney argued against the producer, claiming that the consumption of PeptAid triggered his client's ulcer and resulting medical expenses. Likewise, the plaintiff's attorney argued against the marketer, claiming that his client would not have developed an ulcer, if the marketer had not distributed the product samples.

The defense attorney, representing both the manufacturer and marketer of PeptAid, though, rebutted this argument, stating that the high correlation between PeptAid consumption and ulcers was attributable to a common factor, namely, pre-ulcer discomfort. Individuals with gastrointestinal discomfort would be much more likely to both use PeptAid and develop stomach ulcers. To bolster his clients' claims, the defense attorney introduced expert analysis of the data showing that, on the average, consumption of PeptAid actually decreases an individual's chances of developing ulcers by at least 15%.

Indeed, the application of Eqs. 16 and 17 results in the following bounds on the average causal effect of PeptAid consumption on peptic ulceration

$$-0.23 \leq \text{ACE}(D \rightarrow Y) \leq -0.15$$

and proves that PeptAid is beneficial to the population as a whole.

The plaintiff's attorney, though, stressed the distinction between the average treatment effects for the entire population and the sub-population consisting of those individuals who, like his client, received the PeptAid sample, consumed it and then developed ulcers. Analysis of the population data indicated that had PeptAid not been distributed, Mr. Smith would have had at most a 7% chance of developing ulcers regardless of any confounding factors such as pre-ulcer pain. Likewise, if Mr. Smith had not consumed PeptAid, he would have had at most a 7% chance of developing ulcers.

The damaging statistics against the marketer are obtained by evaluating the bounds on the probability that the plaintiff would have developed a peptic ulcer if he had not received the PeptAid sample, given that he in fact received the sample PeptAid, consumed the PeptAid, and developed peptic ulcers. This probability may be written in terms of the functional model parameters:

$$P(y_1^*|\hat{z}_0^*, y_1, d_1, z_1) = \frac{P(r_z=1)[q_{13} + q_{31} + q_{33}]}{P(y_1, d_1, z_1)}$$

But, since $Z$ is a root node in the probabilistic specification, $P(r_z=1) = P(z_1)$; therefore,

$$P(y_1^*|\hat{z}_0^*, y_1, d_1, z_1) = \frac{q_{13} + q_{31} + q_{33}}{P(y_1, d_1|z_1)}$$
$$= \frac{q_{13} + q_{31} + q_{33}}{p_{11.1}}.$$

This expression is linear with respect to the $Q$ parameters; therefore, we may use linear optimization to derive symbolic bounds on the counterfactual probability with respect to the probabilistic specification $P(y, d|z)$:

$$\frac{1}{p_{11.1}} \max \left\{ \begin{array}{c} 0 \\ p_{11.1} - p_{00.0} \\ p_{11.0} - p_{00.1} - p_{10.1} \\ p_{10.0} - p_{01.1} - p_{10.1} \end{array} \right\}$$
$$\leq P(y_1^*|\hat{z}_0^*, z_1, d_1, y_1) \leq$$
$$\frac{1}{p_{11.1}} \min \left\{ \begin{array}{c} p_{11.1} \\ p_{10.0} + p_{11.0} \\ 1 - p_{00.0} - p_{10.1} \end{array} \right\}$$

Similarly, the damaging evidence against PeptAid's producer is obtained by evaluating the bounds on the counterfactual probability $P(y_1^*|\hat{d}_0^*, y_1, d_1, z_1)$. In terms of the $Q$ parameters the counterfactual probability is written:

$$P(y_1^*|\hat{d}_0^*, y_1, d_1, z_1) = \frac{q_{13} + q_{33}}{q_{11} + q_{13} + q_{31} + q_{33}}$$
$$= \frac{q_{13} + q_{33}}{p_{11.1}}.$$

If we minimize/maximize the numerator given the linear constraints, we arrive at the following bounds:

$$\frac{1}{p_{11.1}} \max \left\{ \begin{array}{c} 0 \\ p_{11.1} - p_{00.0} - p_{11.0} \\ p_{10.0} - p_{01.1} - p_{10.1} \end{array} \right\}$$
$$\leq P(y_1^*|\hat{d}_0^*, z_1, d_1, y_1) \leq$$
$$\frac{1}{p_{11.1}} \min \left\{ \begin{array}{c} p_{11.1} \\ p_{10.0} + p_{11.0} \\ 1 - p_{00.0} - p_{10.1} \end{array} \right\}$$

Substituting the observed distribution $P(y, d|z)$ into these formulas, the following bounds were obtained

$$0.00 \leq P(y_1^*|\hat{z}_0^*, z_1, d_1, y_1) \leq 0.07$$
$$0.00 \leq P(y_1^*|\hat{d}_0^*, z_1, d_1, y_1) \leq 0.07$$

We can write the average causal effects for the subpopulation resembling the plaintiff by conditioning the counterfactual probabilities in Eqs. (10) and (11) on the features of the plaintiff.

$$\text{ACE}(D \to Y|z_1, d_1, y_1) =$$
$$P(y_1^*|\hat{d}_1^*, z_1, d_1, y_1) - P(y_1^*|\hat{d}_0^*, z_1, d_1, y_1)$$

Counterfactual probabilities have the property that if the counterfactual antecedent is implied by the real-world observation, then the probability of the counterfactual consequent is the same as in the real-world given the observations:

$$P(c^*|\hat{a}^*, o) = P(c = c^*|o)$$

Therefore,

$$P(y_1^*|\hat{z}_1^*, z_1, d_1, y_1) = 1.00$$
$$P(y_1^*|\hat{d}_1^*, z_1, d_1, y_1) = 1.00$$

and

$$0.93 \leq \text{ACE}(D \to Y|z_1, d_1, y_1) \leq 1.00$$
$$0.93 \leq \text{ACE}(Z \to Y|z_1, d_1, y_1) \leq 1.00$$

At least 93% of the people in the plaintiff's subpopulation would not have developed ulcers had they not been encouraged to take PeptAid $(z_0)$, or similarly, had they not taken PeptAid $(d_0)$. This lends very strong support for the plaintiff's claim that he was adversely affected by the marketer and producer's actions and product.

The judge ruled in favor of the plaintiff. PeptAid withdrew the product from the market, and initiated a research effort to identify observable characteristics of those individuals who are adversely effected by PeptAid.

# 6 CONCLUSION

This paper has developed a procedure for evaluating bounds on counterfactual probabilities. At first thought, one may believe that assumption-free bounds would be very weak bounds, but this paper has demonstrated that in certain circumstances, the results of such analysis could provide compelling evidence for legal decisions and development of treatment policies.

The corner-stone of counterfactual analysis is the use of functional models with response-function variables, for which the counterfactual probability may be uniquely written. The task of determining bounds involves the optimization of this expression under the constraints imposed by the known probabilistic specification. In general, the task is reduced to the optimization of a polynomial function subject to linear constraints, which introduces the problem of local minima/maxima.

If the counterfactual probability is linear with respect to the functional specification, then the bounds are easily found via linear programming. In addition, in some cases we may be able to derive symbolic bounds on counterfactual probabilities in terms of the probabilistic specification. Such bounds were derived in applications involving: (1) the determination of treatment efficacy from studies where subjects do not comply perfectly with treatment assignment, and (2) the determination of liability in product-safety litigation.

### Acknowledgements

The research was partially supported by Air Force grant #AFOSR 90 0136, NSF grant #IRI-9200918, Northrop Micro grant #92-123, and Rockwell Micro grant #92-122. Alexander Balke was supported by the Fannie and John Hertz Foundation.